\def\year{2018}\relax
\documentclass[letterpaper]{article} 
\usepackage{aaai18}  
\usepackage{times}  
\usepackage{helvet}  
\usepackage{courier}  
\usepackage{url}  
\usepackage{graphicx}  
\usepackage{subfigure}
\begin{document}
%
\title{Exploring Temporal Preservation Networks for Precise Temporal Action Localization}
\author{Ke Yang, Peng Qiao, Dongsheng Li, Shaohe Lv, Yong Dou\\
National Laboratory for Parallel and Distributed Processing,
\\National University of Defense Technology\\
Changsha, China\\
\{yangke13,pengqiao,dongshengli,yongdou,shaohelv\}@nudt.edu.cn\\
}
\maketitle
\begin{abstract}
Temporal action localization is an important task of computer vision. Though a variety of methods have been proposed, it still remains an open question how to predict the temporal boundaries of action segments precisely. Most works use segment-level classifiers to select video segments pre-determined by action proposal or dense sliding windows. However, in order to achieve more precise action boundaries, a temporal
localization system should make dense predictions at a fine granularity.
A newly proposed work exploits Convolutional-Deconvolutional-Convolutional
(CDC) filters to upsample the predictions of 3D ConvNets, making it possible to perform per-frame action predictions and achieving promising performance in terms of temporal action localization. However, CDC network loses temporal information partially due to the temporal downsampling operation. In this paper, we propose an elegant and powerful Temporal Preservation Convolutional (TPC) Network that equips 3D ConvNets with TPC filters. TPC network can fully preserve temporal resolution and downsample the spatial resolution simultaneously, enabling frame-level granularity action localization with minimal loss of time information. TPC network can be trained in an end-to-end manner. Experiment results on public datasets show that TPC network achieves significant improvement on per-frame action prediction and competing results on segment-level temporal action localization.
\end{abstract}

\noindent In recent years, temporal action localization has became a very important
part of computer vision applications. Many works have been proposed to solve
this problem \cite{escorcia2016daps,jiang2014thumos,idrees2017thumos,caba2016fast,rohrbach2012database,oneata2014lear,richard2016temporal,shou2016temporal,singh2016untrimmed,wang2014action,wang2016uts,yeung2016end,shou2017cdc},
but how to perform temporal action localization precisely is still an open question.The purpose of temporal action localization is to determine the boundaries and classes of action segments in untrimmed videos. Most works extract various features on action segments pre-determined by action proposals or sliding windows and use them to train segment-level action classifiers.
Recently, it is claimed that action prediction at a fine granularity is important
for achieving precise action localization \cite{shou2017cdc}. In \cite{shou2017cdc},
a fine-grained action localization framework called Convolutional-De-Convolutional (CDC)
 based on the well-known C3D architecture \cite{tran2015learning} is designed to detect
 actions in every frame. Then frame-level action predictions are used to refine
 the action segment boundaries generated by Segment-CNN (S-CNN) \cite{shou2016temporal}.
 CDC network achieves promising performance in both action predictions at the frame
 granularity and segment-level action localization. However, CDC network loses temporal
 information to some extent since temporal information is compressed during temporal
 downsampling operations. Meanwhile, CDC network's Convolutional-De-Convolutional
filters make two copies of the fully connected (FC) layers of C3D \cite{tran2015learning}
to perform temporal upsampling, resulting in a higher possibility of overfitting.
How can we preserve the temporal length while downsampling the spatial resolution in 3D ConvNets? The most intuitive solution to this problem is reducing the temporal pooling stride to 1. However, this operation changes the temporal receptive field of convolutional filters after the modified pooling layers. This reduces the amount of temporal context that can inform the prediction produced by each unit and also prevents us from using pre-trained models. In order to preserve the temporal receptive field of subsequent layers and take advantage of pre-trained weights rather than train networks from scratch, we replace standard 3D convolutional filters with Temporal Preservation Convolutional (TPC) filters. TPC filters can enlarge the temporal receptive field of standard convolutional filters when using the same kernel size as original convolutional filters. Therefore, TPC can cooperate with pooling layers with a stride of 1 to preserve temporal length of videos and make use of pre-trained weights simultaneously. With TPC, C3D is upgraded to form our TPC network, which can model spatio-temporal information with minimal temporal information loss to make fine-grained action predictions that can be used to refine boundaries of action proposals to precisely localize action segments. Refinement process is shown in Fig. \ref{fig_frm_to_segment}.

It is worth nothing that C3D is designed to label video clips, and needs careful design to conduct frame-level action classification which we believe is important for action localization. The design of temporal preservation architecture, which enables C3D to provide per-frame classification, is non-trivial and needs innovative idea and insight on this task. Although TPC is simple, it equips the convolution layer with the ability to preserve temporal resolution (input temporal length is the same as output length of a conv + pooling stage), no need to perform upsampling with additional layers as CDC.
Our contributions can be concluded as follows: (1) To the best of our knowledge, this is the first work to apply TPC filters, which can fully preserve temporal resolution and downsample spatial resolution simultaneously, allowing network to infer high-level action semantics with no temporal information loss. (2) We apply TPC filters to 3D ConvNets to form TPC networks. Our TPC network can be trained in an end-to-end manner to generate frame-level action predictions which can be used to refine action segments. (3) TPC network achieves promising results in both per-frame action localization and segment-level action localization.

\section{Related Work}
\label{re_work}
\textbf{Action recognition:} Improved Dense Trajectory Feature (iDTF) \cite{wang2011action,wang2013action} consisting of HOG, HOF, MBH features extracted along dense trajectories has been in a dominant position in the field of action recognition. Recently, 2D Convolutional Neural Networks (2DCNN) trained on ImageNet \cite{krizhevsky2012imagenet} to perform RGB image classification such as AlexNet \cite{krizhevsky2012imagenet}, VGG \cite{simonyan2015very}, ResNet \cite{he2016deep} have gradually shown their power, but their performance is limited since they can only capture appearance information. In order to model motion, two-stream ConvNets taking both RBG and optical flow as input have significantly boost the performance \cite{feichtenhofer2016convolutional,wang2016temporal,simonyan2014two}. To model spatio-temporal feature better, 3D CNN architecture called C3D is proposed to extract spatio-temporal abstraction of high-level semantics directly from raw videos \cite{tran2015learning}.

\textbf{Temporal action localization:}
A typical framework used in many state-of-the-art systems \cite{oneata2014lear,singh2016untrimmed,wang2014action,wang2016uts} extracts various features and train a classifier such as Support Vector Machine (SVM) to classify action segments pre-determined by action proposals or densely sliding windows.
Richard and Gall \cite{richard2016temporal} proposed using statistical length and language modeling to represent temporal and contextual structure. Building on techniques for learning sparse dictionaries, \cite{caba2016fast} introduced a sparse learning framework to represent and retrieve action segment proposals of high recall.

In recent years, deep networks improved performance of temporal localization through end-to-end learning from raw video clips directly to localize action segments. A Long Short Term Memory (LSTM)-based agent is trained using REINFORCE to learn  both which frame to look next and when to emit an action segment prediction in \cite{yeung2016end}.
A temporal action proposal framework is designed based on LSTM that takes pre-extracted CNN features in \cite{escorcia2016daps}. In \cite{yeung2015every}, a LSTM network equipped with attention mechanism proposed to model these temporal relations via multiple input and output connections. In \cite{yuan2016temporal}, a Pyramid of Score Distribution Feature (PSDF) capturing the motion information at multiple resolutions centered at each sliding window is proposed and incorporated into the RNN to improve temporal consistency. Sun \emph{et al.} \cite{sun2015temporal} uses web images as prior to train LSTM model to improve action localization performance with only video-level annotations. Although RNN can make use of temporal information to make frame-level prediction, they are usually placed on top of CNN which take a single frame as input rather than directly modeling spatio-temporal abstraction of high-level semantics directly from from raw videos. In addition, RNN based model produces frame-level smoothing that is actually harmful, not beneficial to the task of precise action localization as \cite{yeung2016end} claimed.

Based on C3D \cite{tran2015learning}, an end-to-end Segment-CNN (S-CNN) action localization framework is proposed to improve action localization performance. S-CNN achieves promising results by capturing spatio-temporal information simultaneously. In \cite{shou2017cdc}, a fine-grained action localization framework called Convolutional-De-Convolutional (CDC) is designed to detect actions in every frame. Then frame-level action predictions are used to refine the action segment boundaries generated by S-CNN.

\textbf{Semantic segmentation and atrous convolution:} \cite{chen2014semantic,chen2016deeplab} apply the atrous convolution with upsampled filters dense feature extraction for image segmentation. Atrous convolution allows to explicitly control the resolution at which feature responses are computed within convolutional neural networks. It also allows to effectively enlarge the field of view of filters to incorporate larger context without increasing the number of parameters or the amount of computation. Considering atrous convolution as a powerful tool in dense predict tasks, it shall have the potential to be adapted for making dense predictions in time for our precise temporal action localization task. However, unlike the image segmentation task in which keeping spatial resolution is import, our precise action localization task needs to preserve temporal resolution and downsample spatial resolution simultaneously. To this end, we propose TPC which allows us to preserve temporal resolution  when downsampling spatial resolution at the same time. Our TPC filter can be be regarded as a special case of atrous convolution in the temporal domain.
\begin{figure*}[t]
\centering
\subfigure[C3D's temporal convolution]{
\label{fig_tpc_sub_a}
\includegraphics[height=0.11\textheight]{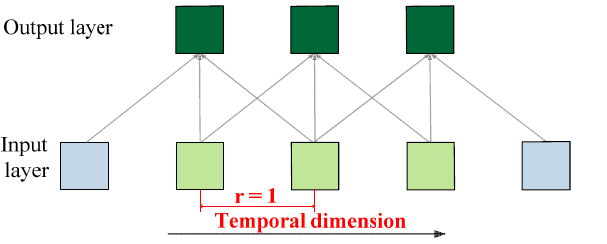}}
\subfigure[temporal preservation convolution]{
\label{fig_tpc_sub_b}
\includegraphics[height=0.11\textheight]{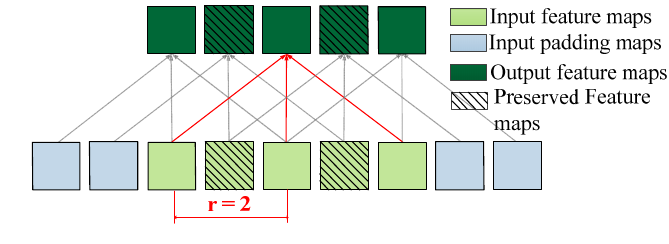}}
\caption{Illustration of temporal preservation convolution. We only show their temporal dimension since spatial dimension is the same. Each box represents the feature maps corresponding to one frame. Bottom line represents input layer while top line represents output layer. (a) Standard temporal convolution on a low resolution feature map that downsampled by pooling layer by a factor of 2. (b) Temporal preservation convolution on a high resolution feature map that is not downsampled. To have the same temporal receptive field size, we need a temporal sample $rate = r$, here $r = 2$.}
\label{fig_tpc}
\end{figure*}

\section{Temporal preservation networks}
\label{TPN}
C3D architecture which consists of five stages 3D ConvNets and three Fully Connected (FC) layers, has been shown that it can learn spatio-temporal patterns from raw video and has promising performance in action recognition \cite{tran2015learning}. However, C3D architecture loses temporal information due to temporal downsampling from conv1a to pool5 layer, and the temporal length of output results in $L/16$ given an input video segment of temporal length $L$. In order to predict actions at a frame-level, CDC network \cite{shou2017cdc} stacks three CDC layers on top of 3D ConvNets part of C3D (3D ConvNets + 3 FCs $\longrightarrow$ 3D ConvNets + 3 CDCs). A CDC filter makes two copies of the fully connected (FC) layers of C3D \footnote{FC layers in C3D have been transformed to convolutional layers following \cite{long2015fully}} to upsample the temporal length by a factor of 2. After temporal upsampling by three times, the temporal length is upsampled to $L$ from $L/8$ \footnote{CDC network keeps temporal length by set pooling stride to 1 in pool5 layer, so its temporal length after pool5 is twice that of C3D} ($L/8 \times 2 \times 2 \times 2 \longrightarrow L$). However, CDC network loses temporal information since it crushes the temporal resolution during the temporal downsampling-upsampling process ($L \rightarrow 8/L \rightarrow L$). In addition, each CDC layer's parameter number is twice that of the corresponding FC layer in C3D, resulting in a higher possibility of overfitting.

In order to make frame-level action predictions without temporal information loss, we had better \emph{preserve temporal resolution throughout the whole forward propagation process rather than using the downsampling-upsampling framework.} To this end, we propose TPC filter and use it to construct a TPC network to make frame-level action predictions.

\subsection{Temporal preservation convolution}
\label{TPN_tpconv}
In this section, we will introduce TPC filter and explain how we build a TPC network with the TPC filters.
Why is temporal resolution reduced in C3D? It has direct relationship with pooling filters whose temporal stride is bigger than 1. To preserve the resolution from beginning to end, we need to reduce all pooling layers' pooling stride to 1. As you will see, we will modify the structure inside 3D ConvNets rather than modify three FC layers as CDC network does. TPC network's operations in spatial dimension are the same as that of C3D, so we mainly consider the temporal dimension next.

As we can see, the modified network can preserve temporal length from beginning to end. However, we can notice that the temporal receptive field \footnote{We name 3D convolutional filters' receptive field's temporal dimension as \emph{temporal receptive field} for convenience} of the convolutional filters after modified pooling layers is smaller than that of standard filters. However, contextual information is very important in disambiguating local cues \cite{galleguillos2010context}.
And this also means we can not use the pre-trained model from C3D, but training a network with a small data set from scratch is very difficult. For these two reasons, we need to increase the convolutional filters' temporal receptive field size to match that of the original convolutional filters.
To this end, we replace the standard 3D convolutional filters in C3D with our TPC filters which can enlarge the temporal receptive field of filters to incorporate larger context without increasing the number of parameters.
Considering only temporal dimension, temporal preservation convolution can be defined as Equation \ref{eq:tpc}, where $x[t]$ \footnote{The shape of $x[t]$ is (number of channels, height, width).} is the feature map corresponding to the $t$-$th$ frame, $w[k]$ is convolutional filter, $K$ is the size of filter, $r$ stands for the stride with which filters sample input. Standard convolution is a special case for stride $r = 1$.
We illustrate TPC in Fig. \ref{fig_tpc}, the convolutional filter samples in previous layer's feature maps' temporal dimension at a stride of 2. TPC filter can also be treated as a bigger filter with fixed zero-value which not updated when network parameters are adjusted. The other parameters are initialized with the pre-trained model and are trainable.

 \begin{equation}\label{eq:tpc}
  y[t] = \sum_{k=1}^{K}  x[t + r \cdot k]w[k]
\end{equation}
The idea of our TPC is similar to that of atrous convolution used in 2D image segmentation \cite{chen2014semantic,chen2016deeplab}, but TPC is performed on temporal dimension rather than spatial dimension. It is worth nothing that directly adaption of atrous convolution to temporal field is non-trivial and needs careful design and experiments and insight on this task. In order to be consistent with \cite{chen2014semantic,chen2016deeplab}, we assign the sampling stride as Temporal Atrous Sampling Rate (TASR). Comparisons of architecture of C3D \cite{tran2015learning}, CDC \cite{shou2017cdc} and our TPC network are shown in Table \ref{table_net_arc}. For C3D, temporal length is downsampled in $pool_{i}$ layers($i = 2,\space3,\space4,\space5$) by a factor of 2 and eventually reduced to $L/16$. CDC network first downsamples temporal resolution to $L/8$ and then stacks three CDC layers to upsample to $L$. Based on C3D, TPC network reduces the pooling stride to 1 in $pool_{i}$ layers($i = 2,\space3,\space4,\space5$), and set $TASR = 2$ for conv3a and conv3b (same as Fig. \ref{fig_tpc_sub_b}), $TASR = 4$ for conv4a and conv4b, and $TASR = 8$ for conv5a and conv5b to keep the temporal length be $L$ from beginning to end.
So TPC network preserves more temporal information than CDC network.
\begin{table*}[t]
\caption{Networks architecture comparison. Illustration of output shape and filter size of each layer. We denote layer-wise output shape using the form of (number of channels $\times$ temporal length $\times$ height $\times$ width). Filter shape using (temporal length$\times$ height $\times$ width, temporal atrous rate) for convolutional layers, and (temporal length$\times$ height $\times$ width, stride (temporal stride, height stride, width stride)) for pooling layers. }
\label{table_net_arc}
\scalebox{10}
\centering
\includegraphics[width=1\textwidth]{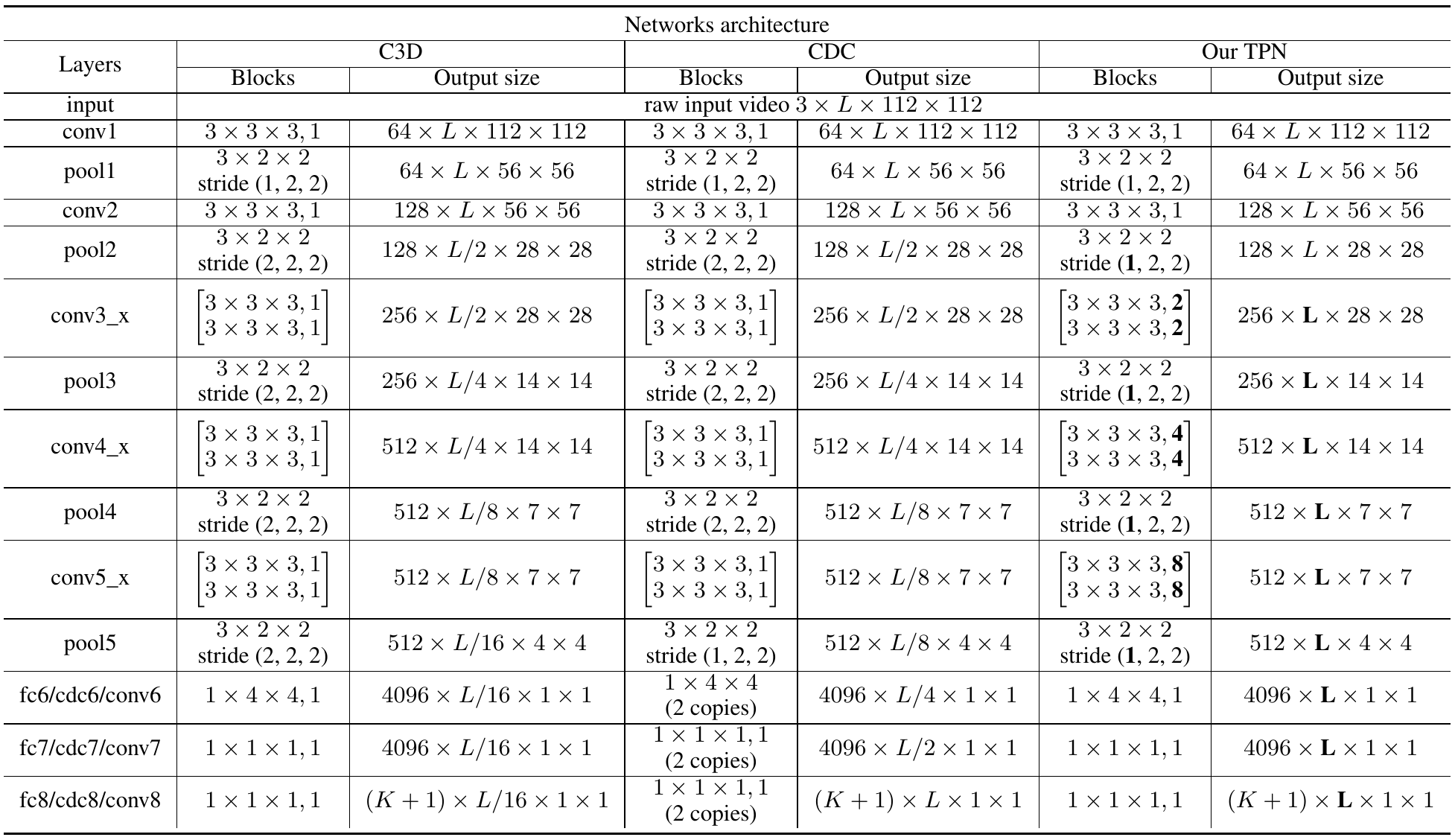}
\end{table*}
\begin{table*}[t]
\caption{Frame-level action localization mAP on THUMOS'14.}
\label{table_frm_level}
\footnotesize
\centering
\begin{tabular}{c|c|c|c|}
\hline
\multicolumn{1}{|c|}{Method} & \multicolumn{1}{c|}{mAP} & \multicolumn{1}{c|}{Method} & \multicolumn{1}{c|}{mAP}\\
\hline
\multicolumn{1}{|c|}{Single-frame CNN\cite{simonyan2015very}} & \multicolumn{1}{c|}{34.7} & \multicolumn{1}{|c|}{TPC-2} & \multicolumn{1}{c|}{45.5} \\
\hline
\multicolumn{1}{|c|}{Two-stream CNN\cite{simonyan2014two}} & \multicolumn{1}{c|}{36.2} & \multicolumn{1}{|c|}{TPC-3} & \multicolumn{1}{c|}{45.1} \\
\hline
\multicolumn{1}{|c|}{LSTM\cite{donahue2015long}} & \multicolumn{1}{c|}{39.3} & \multicolumn{1}{|c|}{TPC-4} & \multicolumn{1}{c|}{45.0} \\
\hline
\multicolumn{1}{|c|}{MultiLSTM\cite{yeung2015every}} & \multicolumn{1}{c|}{41.3} & \multicolumn{1}{|c|}{TPC-2,3} & \multicolumn{1}{c|}{46.4} \\
\hline
\multicolumn{1}{|c|}{Conv \& De-conv\cite{shou2017cdc}} & \multicolumn{1}{c|}{41.7} & \multicolumn{1}{|c|}{TPC-3,4} & \multicolumn{1}{c|}{45.7}\\
\hline
\multicolumn{1}{|c|}{CDC\cite{shou2017cdc}} & \multicolumn{1}{c|}{44.4} & \multicolumn{1}{|c|}{\textbf{TPC}} & \multicolumn{1}{c|}{\textbf{49.5}}\\
\hline
\end{tabular}
\end{table*}

\textbf{More details to construct TPC newtork.} To make it easier to align the output and the input in the temporal dimension, we modify the temporal dimension of all pooling layers' kernel size from 2 to 3. In our descriptions above, details of the convolutional and pooling layers have been clarified. As explained in \cite{long2015fully}, the FC layer is a special case of convolutional layer, and we can transform FC6 (weights shape: $4096 \times 8192$), FC7 (weights shape: $4096 \times 4096$) to conv6 (filter shape: $4096 \times 512 \times 4 \times 4$), conv7 (filter shape: $4096 \times 4096 \times 1 \times 1$) respectively. Now conv6 can slide on $L$ feature maps of size $512\times4\times4$ stacked in time and output $L$ feature maps of size $4096\times1\times1$. Conv6, conv7 layers can be initialized with FC6, FC7, but conv8 can not be adapted from FC8 since output classes are not same in conv8 and FC8,
so we randomly initialize conv8. Following \cite{shou2017cdc}, we perform softmax operation and compute softmax loss for each frame separately. Given a mini-batch with N training segments, batch output $O$ and label $y$, the total loss ${\mathcal{L}}$ is defined as Equation \ref{eq:Loss}. ${\mathcal{L}}$ can be optimized by standard backpropagation (BP)algorithm.

\begin{equation}\label{eq:Loss}
  {\mathcal{L}} = \frac{1}{N} \sum_{n=1}^{N} \sum_{t=1}^{L} \sum_{c=1}^{K + 1} \left(-y^{(c)}_{n}[t]\log \left(\frac{\exp\left(O^{(c)}_{n}[t]\right)}{\sum_{j=1}^{K + 1} \exp\left(O^{(j)}_{n}[t]\right)}\right)\right)
\end{equation}

\subsection{Model training and prediction}
\label{TPN_train}

\textbf{Training data construction.} Training data consists of video segments with length $L$. $L$ can be an arbitrary value because TPC network is a fully convolutional network. We chose $L = 64$ frames in practical due to the Graphics Processing Unit (GPU) memory limitation. Following \cite{shou2017cdc}, we slide temporal window of size $L$ on untrimmed videos and only keep segments include at least one frame belongs to actions to prevent including too many background frames. To construct a balanced training dataset, we re-sample the segments belong to minority classes to ensure each action class has about $80K$ frames.

\textbf{Model training.} We implement TPC network based on Keras \cite{chollet2015keras} and C3D \cite{tran2015learning}. Codes and models will be shared online. We use Stochastic Gradient Descent (SGD) to train TPC network. We first freeze the layers before conv8 and train conv8 with learning rate set to 0.0001, then train all the layer with learning rate set to 0.00001. We set momentum to 0.9 and weight decay to 0.0005. We use C3D \cite{tran2015learning} pre-trained on Sports-1M \cite{karpathy2014large} to initialize TPC network from conv1 to conv7. We randomly initialize weights for conv8.

\textbf{Frame-level action predictions.} During testing, we slide TPC network on the whole video without overlapping. Then, we get the action predictions for all the frames of the whole video. With frame-level features, we can do many things, such as video caption, video action localization. The difference between TPC network frame-level features and 2D CNN frame-level features is that ours are calculated taking into account whole video segment information, so our features are more robust to noise. Compared to 2D CNN+LSTM framework, our frame-level features align more precisely with input since LSTM smooths temporal information \cite{yeung2016end}.

\textbf{Segment-level action predictions.} In order to further verify the effectiveness of TPC network, we carry out segment-level action localization with TPC network's frame-level action predictions. For a direct and fair comparison, first we follow \cite{shou2017cdc} and apply TPC network on proposal segments generated by \cite{shou2016temporal}. We apply the same strategy that using frame-level predictions to refine segment proposals as \cite{shou2017cdc}. We set the category of one segment to the maximum average confidence score over all frames in the video segment. Only the segments not assigned to background class are kept for further boundary refinement. We start from boundaries of each side and move to the middle of the segment, and shrink the temporal boundaries until reach a frame with confidence score lower than the threshold. For more details about the refinement process and the confidence score threshold selecting method please refer to \cite{shou2017cdc}.

In order to make better use of frame-level prediction results, we design a new frame grouping method that gets action segments from untrimmed videos by thresholding on confidence scores and group adjacent frames. First, we take threshold processing on classification scores of all frames in the test video. As a result, we got a string of {"}0{"} and {"}1{"} (0 indicates below the threshold, and 1 inversely). Second, we group the adjacent {"}1{"} to get the segment-level outputs. Then we use NMS to post-process these segments. For threshold value selection, we set multiple different threshold values (uniformly selected from 0 to 1) instead of dataset-dependent. We denote the frame grouping as \textbf{FGM}.

\section{Evaluation}\label{Eval}

We evaluate TPC network on the challenging dataset THUMOS'14 \cite{jiang2014thumos,idrees2017thumos}. Temporal action detection task in THUMOS'14 challenge
is dedicated to localize the action instances in untrimmed video and involves 20 action classes.
Training set consists of 2755 well trimmed videos of these 20 action classes from UCF101 dataset \cite{soomro2012ucf101}. Validation set consists of 1010 untrimmed videos with temporal annotations in form of (video name, action segment start time, action segment ending time, action category). Test set consists of 1574 untrimmed videos. Same as \cite{shou2016temporal,shou2017cdc}, we only keep the videos that contain action instances of interest for testing. We evaluate TPC network on frame-level action localization and segment-level action localization tasks.

\subsection{Frame-level action localization}\label{Eval_frm}

First, we evaluate TPC network in predicting action labels for every frame in the whole video. This task can take multiple frames as input to take into account temporal information.Following \cite{shou2017cdc,yeung2015every}, we evaluate frame-level prediction as a retrieval problem. For each action class, we rank all the images in the test set by their confidence scores and compute Average Precision (AP) for this class. And mean AP (mAP) is computed by average the AP of 20 action classes.

In Table \ref{table_frm_level}, we compare our TPC network with state-of-the-art methods. All the results are quoted from \cite{yeung2015every,shou2017cdc}. Single-frame CNN stands for frame-level VGG-16 2D CNN model in \cite{simonyan2015very}. Two-stream CNN is the frame-level CNN model proposed in \cite{simonyan2014two} using optical flow and RGB images to perform action recognition. LSTM represents the basic 2D CNN + LSTM model proposed in \cite{donahue2015long}. MultiLSTM stands for an extended LSTM using temporal attention mechanism proposed in \cite{yeung2015every}. MultiLSTM uses THUMOS'14 extended version dataset MultiTHUMOS with much more annotations \cite{yeung2015every} to train their network.
Conv \& De-conv stands for the baseline method in \cite{shou2017cdc} replacing CDC layers with de-convolutional layers. CDC stands for the convolutional-de-convolutional network proposed in \cite{shou2017cdc}. We denote our TPC network as \textbf{TPC}. Among these methods, Single-frame CNN only takes into account appearance information in a single frame, Two-stream CNN uses appearance information in a single frame and motion information from two adjacent frames. LSTM and MultiLSTM can make use of temporal information to make frame-level predictions but LSTM based model produces frame-level class probabilities smoothing what is actually harmful, not beneficial to the task of precise action localization as \cite{yeung2016end} claimed. Conv \& De-conv, CDC and our TPC are all based on 3D CNN, can model appearance information and temporal information simultaneously. However, Conv \& De-conv, CDC network both lose temporal information to some extent due to their temporal downsampling process. Our TPC network equipped with TPC filters can perform frame-level predictions with minimal temporal information loss, achieving promising performance.
\begin{table*}[t]
\caption{Segment-level action localization mAP on THUMOS'14. IoU threshold values are ranged from 0.3 to 0.7. '-' in the table indicates that results of that IoU value are not available in the corresponding papers.}
\label{table_segment_level}
\footnotesize
\centering
\begin{tabular}{c|ccccc|c|ccccc}
\multicolumn{1}{c|}{IoU threshold} & \multicolumn{1}{c}{0.3} & \multicolumn{1}{c}{0.4} & \multicolumn{1}{c}{0.5} & \multicolumn{1}{c}{0.6} & \multicolumn{1}{c}{0.7} \\
\hline
\multicolumn{1}{c|}{Wang et al.\cite{wang2014action}} & \multicolumn{1}{c}{14.6} & \multicolumn{1}{c}{12.1} & \multicolumn{1}{c}{8.5} & \multicolumn{1}{c}{4.7} & \multicolumn{1}{c}{1.5} \\
\multicolumn{1}{c|}{Heilbron et al.\cite{caba2016fast}} & \multicolumn{1}{c}{-} & \multicolumn{1}{c}{-} & \multicolumn{1}{c}{13.5} & \multicolumn{1}{c}{-} & \multicolumn{1}{c}{-} \\
\multicolumn{1}{c|}{Escorcia et al.\cite{escorcia2016daps}} & \multicolumn{1}{c}{-} & \multicolumn{1}{c}{-} & \multicolumn{1}{c}{13.9} & \multicolumn{1}{c}{} & \multicolumn{1}{c}{-} \\
\multicolumn{1}{c|}{Oneata et al.\cite{oneata2014lear}} & \multicolumn{1}{c}{28.8} & \multicolumn{1}{c}{21.8} & \multicolumn{1}{c}{15.0} & \multicolumn{1}{c}{8.5} & \multicolumn{1}{c}{3.2} \\
\multicolumn{1}{c|}{Richard and Gall\cite{richard2016temporal}} & \multicolumn{1}{c}{30.0} & \multicolumn{1}{c}{23.2} & \multicolumn{1}{c}{15.2} & \multicolumn{1}{c}{-} & \multicolumn{1}{c}{-} \\
\multicolumn{1}{c|}{Yeung et al.\cite{yeung2016end}} & \multicolumn{1}{c}{36.0} & \multicolumn{1}{c}{26.4} & \multicolumn{1}{c}{17.1} & \multicolumn{1}{c}{-} & \multicolumn{1}{c}{-} \\
\multicolumn{1}{c|}{Yuan et al.\cite{yuan2016temporal}} & \multicolumn{1}{c}{33.6} & \multicolumn{1}{c}{26.1} & \multicolumn{1}{c}{18.8} & \multicolumn{1}{c}{-} & \multicolumn{1}{c}{-} \\
\multicolumn{1}{c|}{S-CNN\cite{shou2016temporal}} & \multicolumn{1}{c}{36.3} & \multicolumn{1}{c}{28.7} & \multicolumn{1}{c}{19.0} & \multicolumn{1}{c}{10.3} & \multicolumn{1}{c}{5.3} \\
\multicolumn{1}{c|}{Conv \& De-conv\cite{shou2017cdc} + S-CNN\cite{shou2016temporal}} & \multicolumn{1}{c}{38.6} & \multicolumn{1}{c}{28.2} & \multicolumn{1}{c}{22.4} & \multicolumn{1}{c}{12.0} & \multicolumn{1}{c}{7.5} \\
\multicolumn{1}{c|}{CDC\cite{shou2017cdc} + S-CNN\cite{shou2016temporal}} & \multicolumn{1}{c}{40.1} & \multicolumn{1}{c}{29.4} & \multicolumn{1}{c}{23.3} & \multicolumn{1}{c}{13.1} & \multicolumn{1}{c}{7.9} \\
\hline
\multicolumn{1}{c|}{TPC-2 + S-CNN\cite{shou2016temporal}} & \multicolumn{1}{c}{37.8} & \multicolumn{1}{c}{28.9} & \multicolumn{1}{c}{22.6} & \multicolumn{1}{c}{13.7} & \multicolumn{1}{c}{7.8} \\
\multicolumn{1}{c|}{TPC-3 + S-CNN\cite{shou2016temporal}} & \multicolumn{1}{c}{37.6} & \multicolumn{1}{c}{29.0} & \multicolumn{1}{c}{22.3} & \multicolumn{1}{c}{13.3} & \multicolumn{1}{c}{7.4} \\
\multicolumn{1}{c|}{TPC-4 + S-CNN\cite{shou2016temporal}} & \multicolumn{1}{c}{37.6} & \multicolumn{1}{c}{28.7} & \multicolumn{1}{c}{22.1} & \multicolumn{1}{c}{12.7} & \multicolumn{1}{c}{6.9} \\
\multicolumn{1}{c|}{TPC-2,3 + S-CNN\cite{shou2016temporal}} & \multicolumn{1}{c}{39.8} & \multicolumn{1}{c}{30.7} & \multicolumn{1}{c}{24.1} & \multicolumn{1}{c}{13.9} & \multicolumn{1}{c}{7.8} \\
\multicolumn{1}{c|}{TPC-3,4 + S-CNN\cite{shou2016temporal}} & \multicolumn{1}{c}{38.5} & \multicolumn{1}{c}{29.3} & \multicolumn{1}{c}{22.9} & \multicolumn{1}{c}{13.5} & \multicolumn{1}{c}{7.6} \\
\multicolumn{1}{c|}{\textbf{TPC} + S-CNN\cite{shou2016temporal}} & \multicolumn{1}{c}{\textbf{41.9}} & \multicolumn{1}{c}{\textbf{32.5}} & \multicolumn{1}{c}{\textbf{25.3}} & \multicolumn{1}{c}{\textbf{14.7}} & \multicolumn{1}{c}{\textbf{9.0}} \\
\hline
\multicolumn{1}{c|}{CDC\cite{shou2017cdc} + FGM} & \multicolumn{1}{c}{36.1} & \multicolumn{1}{c}{28.2} & \multicolumn{1}{c}{20.9} & \multicolumn{1}{c}{14.9} & \multicolumn{1}{c}{8.1} \\
\multicolumn{1}{c|}{\textbf{TPC} + FGM} & \multicolumn{1}{c}{\textbf{44.1}} & \multicolumn{1}{c}{\textbf{37.1}} & \multicolumn{1}{c}{\textbf{28.2}} & \multicolumn{1}{c}{\textbf{20.6}} & \multicolumn{1}{c}{\textbf{12.7}} \\
\end{tabular}
\end{table*}
In addition, in order to verify the effectiveness of TPC on temporal information preservation, we compare TPC with TPC's variants that only use TPC filters on one or two layers. (1) TPC-2: we only use TPC in conv2. (2) TPC-3: we only use TPC in conv3. (3) TPC-4: we only use TPC in conv4. (4) TPC-2,3: we use TPC in conv2 and conv3. (5) TPC-3,4: we use TPC in conv3 and conv4. Complete TPC network use TPC filters on conv2, conv3 and conv4 (i.e., TPC-2,3,4). For the five variants, we apply linear interpolation to upsample predictions to output frame-level predictions for both training and testing. We train them using the same training data as TPC.
Comparisons suggest that preserving temporal information at early stage helps preserve more details and brings better result, but not that much. TPC-2,3,4 brings notable performance improvement, suggesting that preserving the temporal resolution in all layers brings minimal temporal information loss and better performance.

\begin{figure*}[t]
\centering
\includegraphics[width=1\textwidth]{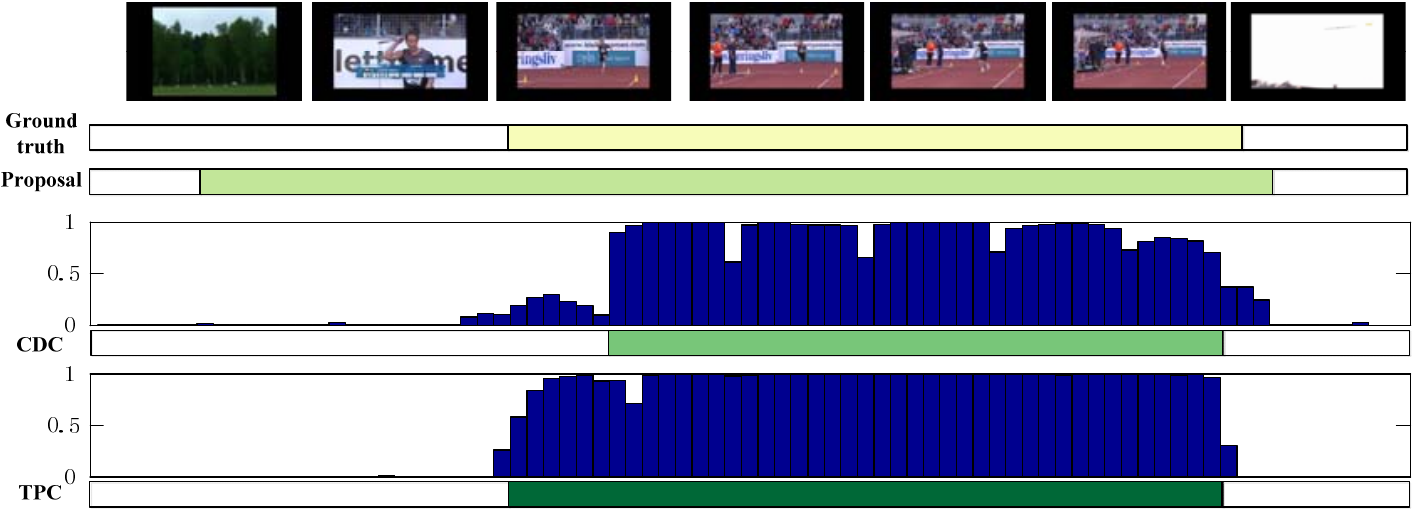}
\caption{Illustration of the process of temporal boundaries refinement using frame-level predictions. Horizontal axis stands for time and vertical axis stands for confidence score. From the top to the bottom: (1) frame-level ground truth for a JavelinThrow instance in an input video; (2) corresponding proposal generated from \cite{shou2016temporal}; (3) frame-level predictions of CDC \cite{shou2017cdc} and refined action instance using CDC; (4) frame-level predictions of TPC and refined action instance using TPC.}
\label{fig_frm_to_segment}
\end{figure*}

\subsection{Temporal action localization}\label{Eval_segment}

Given frame-level action predictions, we can get segment-level action localization results using various strategies. For more direct comparison, we first use the same strategy as CDC \cite{shou2017cdc}. First, we generate action segment proposals using the S-CNN\cite{shou2016temporal}; second, each segment is set to an action category; then, non-background segments' boundaries are refined with frame-level action predictions and confidence scores are calculated by averaging confidence scores of all the frame in refined segments; finally, we perform post-processing steps such as non-maximus suppression.
We evaluate our model on THUMOS'14 dataset.

We perform evaluation using mAP as frame-level action localization evaluation. For each action class, we rank all the predicted segments by their confidence results and calculate the AP using official evaluation code. One prediction is correct when its temporal overlap intersection-over-union (IoU) with a ground truth action segment is higher than the threshold, so evaluation under various IoU threshold is necessary. We evaluate our model under IoU threshold from 0.3 to 0.7. Results are shown in Table \ref{table_segment_level}, our model denoted as \textbf{TPC} achieves better results than other methods.

As shown in Table \ref{table_frm_level} and Table \ref{table_segment_level}, TPC achieves clearly improvement over other baselines on frame-level task but the improvement is far less significant on segment-level task.
The reason might be that proposals by S-CNN\cite{shou2016temporal} help CDC\cite{shou2017cdc} much more. Proposals from \cite{shou2016temporal} help CDC or TPC filter video segments which might be background frames. TPC performs much better than CDC on frame-level task, which means that TPC also does much better on the filtered frames. So proposals do not improve TPC’s performance that much as CDC. To verify this idea, we perform FGM on both TPC and CDC frame-level classification results to get segment-level detections. Results are shown in Table \ref{table_segment_level}, TPC’s performance improves significantly after using the new frame grouping method. The reason for the significant improvement is that proposals from \cite{shou2016temporal} have false negatives, and TPC can handle these false negative frames. CDC’s\cite{shou2017cdc} performance decrease (when IoU = 0.3, 0.4, 0.5) because their inferior performance outside the proposals. Overall,results suggest that frame-level results indeed contributes to precise segment-level localization.

Quantitative experiment results are shown in Fig. \ref{fig_frm_to_segment}. This results suggest that TPC perform better on frame-level classification, and this better results lead to better segment-level results. We also can clearly observe that CDC suffered from checkerboard artifacts brought by the deconvolution operations \cite{odena2016deconvolution}. Our TPC is not affected by this problem because TPC can preserve temporal length and does not need to use deconvolution to upsample in time.

\subsection{Discussion}\label{Eval_disc}
TPC network allows us to compute feature responses at the original video temporal resolution, but it indeed increases computational overhead. In order to give a fair comparison, we implemented CDC network \cite{shou2017cdc} in our experiment environments. On a NVIDIA Titan X GPU with 12GB memory, our TPC can predict around 250 frames per second (FPS) while CDC network predicts around 390 FPS. Although our method is not as fast as CDC network, it is enough for real-time application. After all, our TPC network can process 10 seconds video clip of 25 FPS within one second.

We also try another variant of TPC network that we add global average pooling (GAP) layer on pool5 layer of TPC network and then add a conv6-GAP layer to output K + 1 classes confidence scores (using suffix -GAP to distinguish with conv6 layer in original TPC network). We denote this variant as TPC-GAP. TPC-GAP network achieves 47.2 mAP in frame-level action localization and 23.6 mAP with 0.5 IoU threshold in segment-level action localization. TPC-GAP has only 1/5 of CDC network's parameter but can achieve competitive results.

\section{Conclusion}

In this paper, we propose a TPC filter to replace the standard convolutional filters in 3D ConvNets. Then we use TPC filters to construct our TPC network. Our TPC network can make more precise frame-level action predictions since it preserve all the temporal information. We also evaluate our model on segment-level action localization task. Experiments on frame-level and segment-level action localization tasks both suggest that our model achieves superior results compared with previous works. TPC network can predict around 250 frames per second which is good news for real-time applications. In addition, our TPC filter can be adapted for other applications, such as combined with the
spatial atrous convolutional filter to perform video segmentation.

\bibliography{aaai2018}
\bibliographystyle{aaai}
\end{document}